\documentclass[conference]{IEEEtran}
\IEEEoverridecommandlockouts

\usepackage{times,booktabs,tabularx, bm, array,subcaption,hyperref,multirow,pifont}
\hypersetup{
    colorlinks=true,
    linkcolor=blue,
    filecolor=magenta,
    urlcolor=cyan,
    pdftitle={Overleaf Example},
    pdfpagemode=FullScreen,
    }
\usepackage{algorithmic}
\usepackage{graphicx}
\usepackage{textcomp}
\usepackage{xcolor}
\usepackage{cite}
\usepackage{amsmath,amssymb,amsfonts}
\usepackage{algorithmic}
\usepackage{graphicx}
\usepackage{textcomp}
\usepackage{xcolor}

\def\BibTeX{{\rm B\kern-.05em{\sc i\kern-.025em b}\kern-.08em
    T\kern-.1667em\lower.7ex\hbox{E}\kern-.125emX}}

\title{Band Prompting Aided SAR and Multi-Spectral Data Fusion Framework for Local Climate Zone Classification}

\begin{document}

\author{
    \IEEEauthorblockN{
        Haiyan Lan$^{1}$, 
        Shujun Li$^{1}$, 
        Mingjie Xie$^{2}$,
        Xuanjia Zhao$^{1}$,
        Hongning Liu$^{3}$,\\
        Pengming Feng$^{4}$,
        Dongli Xu$^{5}$,
        Guangjun He$^{4}$,
        Jian Guan$^{1*}$\thanks{*~Corresponding author.}
    }
    \IEEEauthorblockA{
        $^{1}$College of Computer Science and Technology, Harbin Engineering University, Harbin, China\\
        $^{2}$School of Astronautics, Beihang University, Beijing, China\\
        $^{3}$School of Software, Dalian University of Technology, Dalian, China\\
        $^{4}$State Key Laboratory of Space-Ground Integrated Information Technology, Beijing, China\\
        $^{5}$Independent Researcher\\
    }
}

\maketitle

\begin{abstract}
Local climate zone (LCZ) classification is of great value for understanding the complex interactions between urban development and local climate. 
Recent studies have increasingly focused on the fusion of synthetic aperture radar (SAR) and multi-spectral data to improve LCZ classification performance. However, it remains challenging due to the distinct physical properties of these two types of data and the absence of effective fusion guidance.
In this paper, a novel band prompting aided data fusion framework is proposed for LCZ classification, namely BP-LCZ, which utilizes textual prompts associated with band groups to guide the model in learning the physical attributes of different bands and semantics of various categories inherent in SAR and multi-spectral data to augment the fused feature, thus enhancing LCZ classification performance.
Specifically, a band group prompting (BGP) strategy is introduced to align the visual representation effectively at the level of band groups, which also facilitates a more adequate extraction of semantic information of different bands with textual information.
In addition, a multivariate supervised matrix (MSM) based training strategy is proposed to alleviate the problem of positive and negative sample confusion by completing the supervised information.
The experimental results demonstrate the effectiveness and superiority of the proposed data fusion framework.

\end{abstract}


\section{Introduction}
\label{sec:intro}
Local climate zone (LCZ) classification plays a crucial role in urban environment research such as urban heat island (UHI) and urban planning, by providing an objective and culturally agnostic classification scheme of urban and natural landscapes \cite{yang2020optimizing}.
According to the surface structure, surface cover and human activities, LCZ scheme introduces 17 classes, including 10 built types and 7 land cover types \cite{huang2023mapping}. The standardized taxonomy provided by LCZ scheme has led to its increasing application across a wide range of urban-related studies, such as population density estimation, disaster mitigation and infrastructure planning \cite{perera2018local, LCZuse4}.

Recently, several studies \cite{feng2019embranchment, fusion1, fusion2, dynamic} make efforts to fuse the multimodal remote sensing data, e.g., SAR and multi-spectral, to enhance LCZ classification performance by exploiting their complementary information \cite{mazza2023synergic}.
Among them, both \cite{feng2019embranchment} and \cite{ fusion2} employ band grouping techniques to facilitate the fusion of multimodal features.
However, the lack of effective guidance in these approaches prevents the adequately capture of the distinct physical properties of each modality, limiting the ultimate classification performance. 
Interestingly, recent studies in visual representation, such as \cite{multimodalforvisualrepresentation} and \cite{mvp}, can address a similar problem of limited visual feature.
These methods, which introduce text data with visual data, have shown promising results in visual recognition and classification tasks.

\begin{figure*}[htbp]
    \centering
    \includegraphics[width=0.75\textwidth]
    {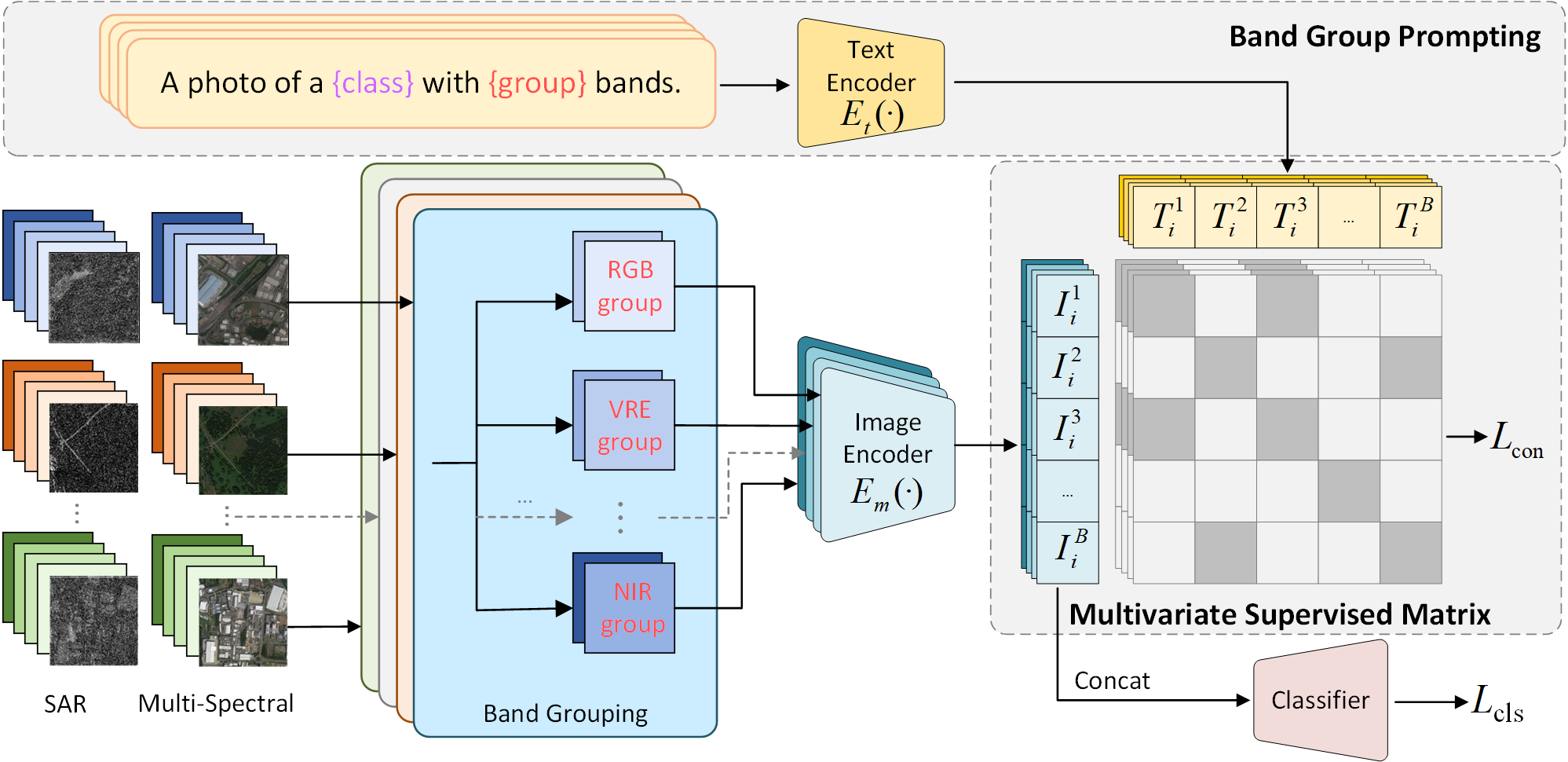}
    \caption{The overall framework of the proposed method, which consists of four steps, i.e., band grouping, band group prompting (BGP), multivariate supervised matrix (MSM) and classification. BGP is used to align image features and text features during training, so as to guide the model to learn the physical attributes of different bands and semantics of different categories contained in SAR and multi-spectral data to augment the feature fusion. MSM is used to alleviate the positive and negative sample confusion problem.
}
    \label{fig:overall}
    \vspace{-5mm}
\end{figure*}

Building on recent advancements, this paper explores the use of textual prompts to aid in multimodel remote sensing data~(i.e., SAR and multi-spectral) fusion.
Inspired by \cite{feng2019embranchment} and \cite{radford2021learning}, we first divide multiple bands of SAR and multi-spectral data into different band groups based on their physical characteristics, while each group is processed separately by dedicated image encoders for better band information extraction. Then,  the band and semantic information are introduced via textual prompts to guide the extraction of visual features from different bands and modalities, enabling the fused features that can reflect band characteristics and semantic content accurately. Specifically, a band group prompting (BGP) strategy is proposed, which generates customized textual prompts for each band group and aligns textual prompts and band groups within a unified feature space. It allows the model to extract visual representation guided by band-specific and semantic information, using textual prompts associated with band groups and LCZ categories.

In addition, following \cite{radford2021learning}, we jointly train the image encoder and text encoder to predict the correct pairings of a batch of input image-text training samples. However, \cite{radford2021learning} only treats the diagonal elements of the similarity matrix as positive samples and others as negative samples, due to the image-text pairs used in the \cite{radford2021learning} are one-to-one correspondence, which is suboptimal for the LCZ classification task with customized textual prompts for each band group. Therefore, a multivariate supervised matrix (MSM) based training strategy is introduced in our study to alleviate the problem of positive and negative sample confusion.

In summary, this paper proposed a band prompting aied SAR and multi-spectral data fusion framework for LCZ classification, namely BP-LCZ, which consists of two main contributions, i.e., BGP strategy and MSM based training strategy. Specifically, BP-LCZ first utilizes BGP strategy to guide the model in learning the physical attributes of different bands and semantics of various categories inherent in SAR and multi-spectral data to augment the fused feature. Then, MSM based training strategy is introduced to complete the supervised information, thus enhancing LCZ classification performance.

\section{Methodology}
\label{sec:method}

To utilize and fuse the information of different bands in SAR and multi-spectral data, we introduce a band prompting aided data fusion framework for LCZ classification. The overall framework of our proposed method is illustrated in Fig. \ref{fig:overall}, which is mainly composed of four steps, i.e., band grouping, band group prompting strategy, multivariate supervised matrix based training strategy and classification.

\begin{table*}[!htbp]
\newcommand{\tabincell}[2]{\begin{tabular}{@{}#1@{}}#2\end{tabular}}
\centering
\caption{Examples of the extended description used to replace the LCZ class name in prompt template. Here, only three LCZ categories are given to illustrate that the extended descriptions have more detailed category information.}
\label{tab:des}
\vspace{+1mm}
\setlength{\tabcolsep}{2mm}{
\resizebox{\linewidth}{!}{
\begin{tabular}{lll}
\toprule
ID & Class Name & Extended Description\\
\midrule
1 & compact high-rise & \textit{dense clusters of vertical edifices with limited spatial gaps, representing urban centers}\\
9 & sparsely built & \textit{areas with minimal building density, characterized by isolated structures and significant open spaces}\\
G & water & \textit{bodies of water such as rivers, lakes, or seas, appearing as uniform blue areas on images}\\
\bottomrule
\end{tabular}}}
\vspace{-5mm}
\end{table*}

\subsection{Band Grouping}
\label{ssec:bg}
In order to better utilize the rich band information in SAR and multi-spectral data, our approach first separates the data into different band groups based on the physical properties of the different bands. Following \cite{feng2019embranchment} and \cite{group},  we also divide SAR and multi-spectral data into 7 different groups according the spectral characteristics of different bands. Specifically, SAR data is divided into 3 band groups, i.e., 1st, 2nd and 5th bands for VH group, 3rd, 4th and 6th bands for VV group, and 7th and 8th bands for PolSAR group. As for multi-spectral data, it is divided into 4 band groups, i.e., 1st, 2nd and 3rd band for RGB group, 4th, 5th, 6th and 8th bands for VRE group, 7th band fo NIR group, and 9th and 10th band for SWIR group. 
Let $\ddot{\bf{X}}$ and $\tilde{\bf{X}} \in \mathbb{R}^{H \times W \times C}$ denote the input SAR and multi-spectral data, respectively. Here, $H$, $W$ and $C$ represent the height, width and channels of the input data, respectively. After band grouping, the input SAR and multi-spectral data are divided into the set of band groups $\{{\bf{X}}_i\in\mathbb{R}^{H \times W \times C_i}\}_{i=1}^n$, where $n$ means the total number of band groups, i.e., $n=7$ in this work. ${\bf{X}}_i$ indicates the $i$-th band group.  

\subsection{Band Group Prompting Strategy}
\label{ssec:bgp}
\subsubsection{Prompt Design for Band Groups}
We generate textual prompt ${\bf{t}}_i$ for each band group ${\bf{X}}_i$ by using specially designed prompt template, which can be formalized as ``a photo of a \textit{\{class\}} with \textit{\{group\}} bands". Here,  \textit{\{class\}} indicates the category information, and \textit{\{group\}} represents the band group information. In this paper, the category information is described empirically in an extended description instead of the simple class name or index, as shown in Table \ref{tab:des}. Whereas the band group information is described with the name of the band grouping, e.g., using ``red green blue'', ``vegetation red edge'', ``near infrared'', ``short wave infrared'', ``vh'', ``vv'' and ``pol'' to represent the RGB, VRE, NIR, SWIR, VH, VV, and PolSAR band groups, respectively. Specifically, taking the RGB band group with the category of water as an example, the generated textual prompt is ``{\bf a photo of a \textit{bodies of water such as rivers, lakes, or seas, appearing as uniform blue areas on images} with \textit{red green blue} bands}". In this way, the prompts for different band groups are generated.

\subsubsection{Feature Extraction}
\label{ssec:feature}
For each band group, we employ the image-text dual-encoder scheme to obtain the features of the two modalities separately. Specifically, given a band group ${\bf{X}}_i$ and the corresponding textual prompt ${\bf{t}}_i$, the image feature ${\bf{I}}_i$ and text feature ${\bf{T}}_i$ are extracted as follows:
\begin{equation}
{\bf{I}}_i = E_m({\bf{X}}_i)
\end{equation}
\begin{equation}
{\bf{T}}_i = E_t({\bf{t}}_i)
\end{equation} 
where $E_m(\cdot)$ and $E_t(\cdot)$ represent the image encoder and text encoder, respectively. Thus, the set of image features $\mathcal{I}_i=[{\bf{I}}_i^1,{\bf{I}}_i^2,\cdots,{\bf{I}}_i^B]$ and text features $\mathcal{T}_i=[{\bf{T}}_i^1,{\bf{T}}_i^2,\cdots,{\bf{T}}_i^B]$ for the $i$-th band group among a batch of input data are obtained. Here, $B$ represents the batch size.
After that, $\mathcal{I}_i$ and $\mathcal{T}_i$ are normalized and then multiplied directly to calculate the cosine similarity to obtain image-text similarity matrix ${\bf S}_i\in\mathbb{R}^{B\times B}$ and text-image similarity matrix $\hat{\bf S}_i\in\mathbb{R}^{B\times B}$. Then, we weight and fuse the similarity matrices of each band group to obtain the final similarity matrices ${\bf S}$ and $\hat{\bf S}$ for model training as
\begin{equation}
{\bf S} = \alpha \sum_i^n {\bf S}_i
\end{equation}
\begin{equation}
\hat{\bf S} = \alpha \sum_i^n \hat{\bf S}_i
\end{equation}
where ${\alpha}$ denotes the weighting coefficient.

\subsection{Multivariate Supervised Matrix Based Training Strategy}
\label{ssec:msm} 
Considering that the categories in a batch of data may be the same and actual positive sample pairs may appear in the off-diagonal position of the similarity matrix, the multivariate supervised matrix (MSM) based training strategy is proposed to complete the real positive samples in the supervised matrix, thereby enhancing the model's performance.

Specifically, we first construct the supervised matrix ${\bf W} \in \mathbb R^{B\times B}$ based on the categories of the training data for each batch, as follows:
\begin{equation}
{\bf W}_{j,k} = \begin{cases}
1, \quad & \text{if} \quad y^*_j=y^*_k \\
0, \quad & \text{otherwise}
\end{cases} 
\end{equation}
where ${y^*}$ denotes the label of the sample, and the range of ${j}$ and ${k}$ is $[1, B]$. Here, the positions with the value of 1 in the supervised matrix are considered as positive samples and other positions are considered as negative samples. 
After obtaining the supervised matrix ${\bf W}$ and the similarity matrices ${\bf S}$ and $\hat{\bf S}$, the image-text based contrastive learning loss function $L_{con}$ is formulated, as follows:
\begin{equation}
L_{con} = (L_{i2t}+L_{t2i})/2
\end{equation}
where $L_{i2t}$ and $L_{t2i}$ denote the image-text and text-image loss functions  respectively, as follows:
\begin{equation}
L_{i2t} = BCE(\sigma({\bf S}),{\bf W})
\end{equation}
\begin{equation}
L_{t2i} = BCE(\sigma(\hat{\bf S}),{\bf W})
\end{equation}
where $BCE(\cdot)$ represents binary cross-entropy loss function, and $\sigma(\cdot)$ is the Sigmoid function.

\subsection{Classification}
\label{ssec:classify}
After acquiring the image features for each band group, our approach concatenates them for feature fusion and feeds them into the classifier ${\bf \mathcal{D}}$. The fused feature ${\bf I}$ is obtained as follows:
\begin{equation}
{\bf I} = Concat([{\bf I}_{1}, {\bf I}_{2}, \cdots , {\bf I}_{n}])
\end{equation}
Thus, we can obtain the predicted labels as $\hat{{\bf c}}={\bf \mathcal{D}}({\bf I})$. Then, the total loss function $L$ is constructed with two types of loss, i.e, classification loss $ L_{cls}$ and contrastive learning loss $L_{con}$, denoted as:
\begin{equation}
L = L_{cls} + \beta * L_{con}
\end{equation}
where ${\beta}$ is the penalty parameter to balance these two types of loss. The classification loss is formulated as:
\begin{equation}
L_{cls} = CE(\hat{{\bf c}},{\bf c^*})
\end{equation}
where $CE(\cdot)$ denotes the cross-entroy loss function, and ${\bf c^*}$ is the ground truth for classification.

\section{Experiments}
\label{sec:exp}

\subsection{Experimental Setup}
\label{ssec:exp_details}
\noindent\textbf{\textit{Dataset:}} The So2Sat LCZ42 dataset \cite{zhu2019so2sat} is adopted for performance evaluation, which contains paired SAR and multi-spectral image patches from Sentinel-1 and Sentinel-2 satellites, respectively, with 17 LCZ categories. All the image patches are 32$\times$32 pixels, whereas the number of channels for SAR and multi-spectral image patches are 8 and 10, respectively. In this dataset, images are collected from different cities around the world. Thus, in order to avoid category imbalance and domain shift due to geographic bias, we reorganized the dataset for the experiments from the training set. Specifically, following \cite{feng2019embranchment}, we randomly selected 1306 samples for each category to construct training set and 12117 samples as the test set, and there are no intersections between the training and test sets.

\noindent\textbf{\textit{Implementation:}} 
In our experiments, the models are trained for 200 epochs, where the learning rate is set to be 0.0001 and the batch size is 32. 
Stochastic gradient descent~(SGD) \cite{SGD} is used as the optimizer, and the momentum and weight decay are set to be 0.9 and 0.002, respectively. The hyperparameters ${\alpha}$ and ${\beta}$ are set to 0.25 and 2. For ease of reproduction, the random seed is fixed at 47.
Overall accuracy (OA) and Kappa coefficient are used as the evaluation metrics following \cite{group}.

\subsection{Results}
\label{ssec:results}
\subsubsection{Performance Comparison}
\label{sssec:comparison}
To validate BP-LCZ for feature fusion to enhance LCZ classification performance, we adopt it for performance comparison with ResNet \cite{resnet}, ResNeXt \cite{ResNeXt}, ConvNeXt \cite{convnext}, DenseNet121 \cite{huang2017densely46}, ExViT \cite{yao2023extended21} and EB-CNN \cite{feng2019embranchment}. Among these methods, ResNet, ResNeXt, DenseNet121 and ConvNeXt are methods commonly used in natural images, EB-CNN and ExViT are methods used in remote sensing images. Our BP-LCZ is adopted in EB-CNN and ExViT to show its advantages, denoted as EB-CNN (BP-LCZ) and ExViT (BP-LCZ), respectively.

\begin{table}[htbp]
  \centering
  \caption{Performance comparison with the different models.}
  \vspace{+1mm}
  \begin{tabular}{lcc}
    \toprule
    Method & OA (\%) & Kappa (\%)\\
    \midrule
    ResNet50 \cite{resnet} & 51.14 & 47.10 \\
    ResNeXt  \cite{ResNeXt} & 53.68 & 49.69 \\
    ConvNeXt \cite{convnext} & 62.84 & 59.49 \\
    DenseNet121 \cite{huang2017densely46} & 64.74 & 61.68 \\
    \midrule
    ExViT  \cite{yao2023extended21} & 77.21 & 75.10 \\
    \textbf{ExViT (BP-LCZ)}  & 79.85 (+2.64) & 77.91 (+2.81) \\
    \midrule
    EB-CNN \cite{feng2019embranchment} & 79.29 & 77.39 \\
    \textbf{EB-CNN (BP-LCZ)} & \textbf{86.69 (+7.40)} & \textbf{85.40 (+8.01)} \\ 
    \bottomrule
  \end{tabular}
  \label{tab:prf}
  \vspace{-3mm}
\end{table}

As illustrated in Table \ref{tab:prf}, OA and Kappa coefficient of ResNet50, ResNeXt, DenseNet121 and ConvNeXt are notably lower than those of ExVit and EB-CNN, indicating that models designed for natural images are not well-suited for remote sensing image analysis. By integrating the BP-LCZ strategy into ExViT and EB-CNN, we observed a substantial improvement, with OA increasing by 2.64\% and 7.40\%, and the Kappa coefficient improving by 2.81\% and 8.01\%, respectively. 
These results demonstrate that BP-LCZ can effectively achieve the fusion of SAR and multi-spectral data through band grouping, and further refine visual representations using textual prompts, thereby enhancing feature fusion and overall performance.
\begin{figure}[htbp]
    \centering
    \vspace{-4mm}
    \includegraphics[width=.85\linewidth]{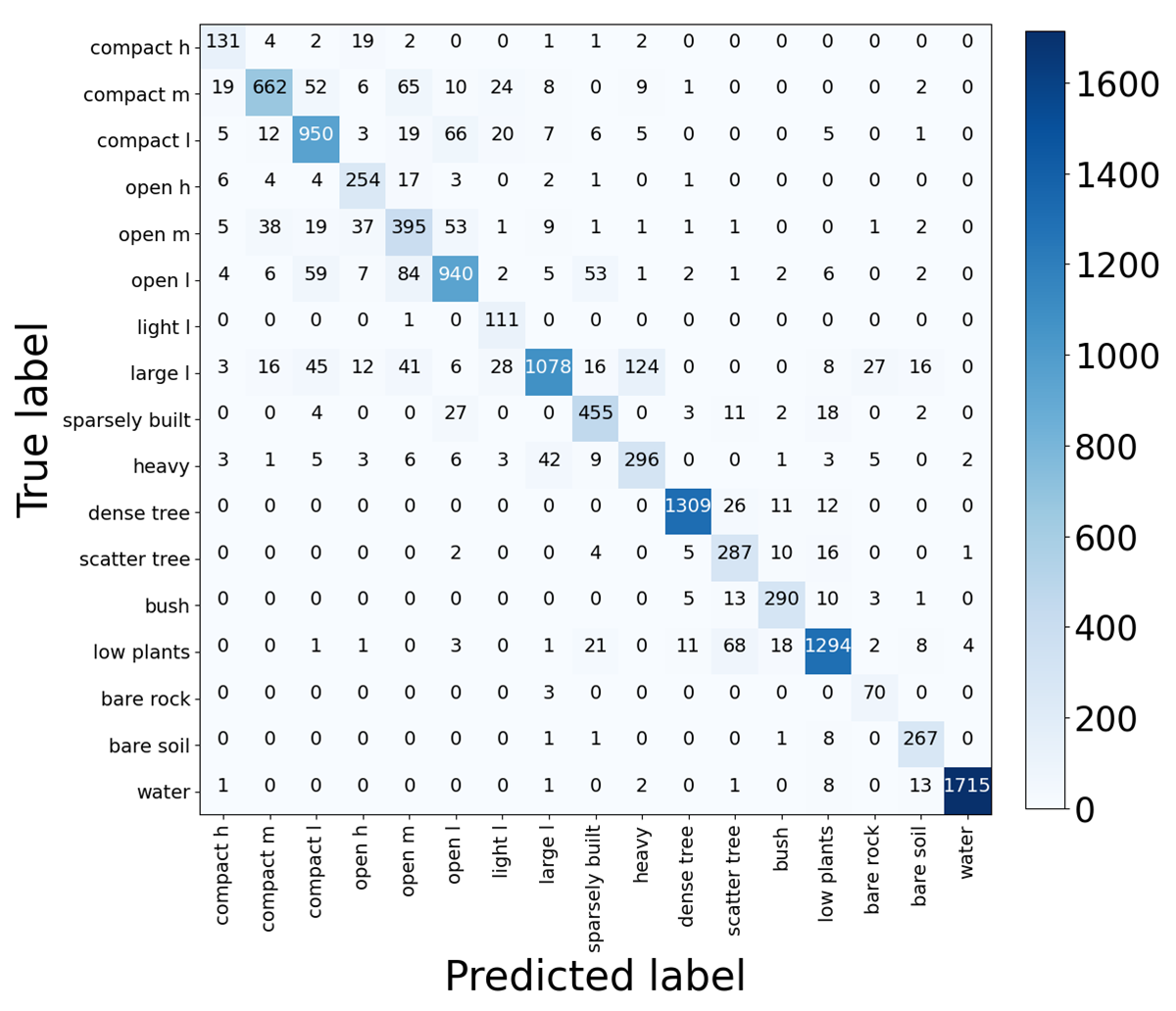}
    \caption{Confusion matrix of the classification results from EB-CNN with the proposed BP-LCZ method.}
    \label{fig:confusion}
   \vspace{-2mm}
\end{figure}

\begin{figure}[htbp]
    \centering
    \vspace{-1mm}
    \includegraphics[width=.75\linewidth]{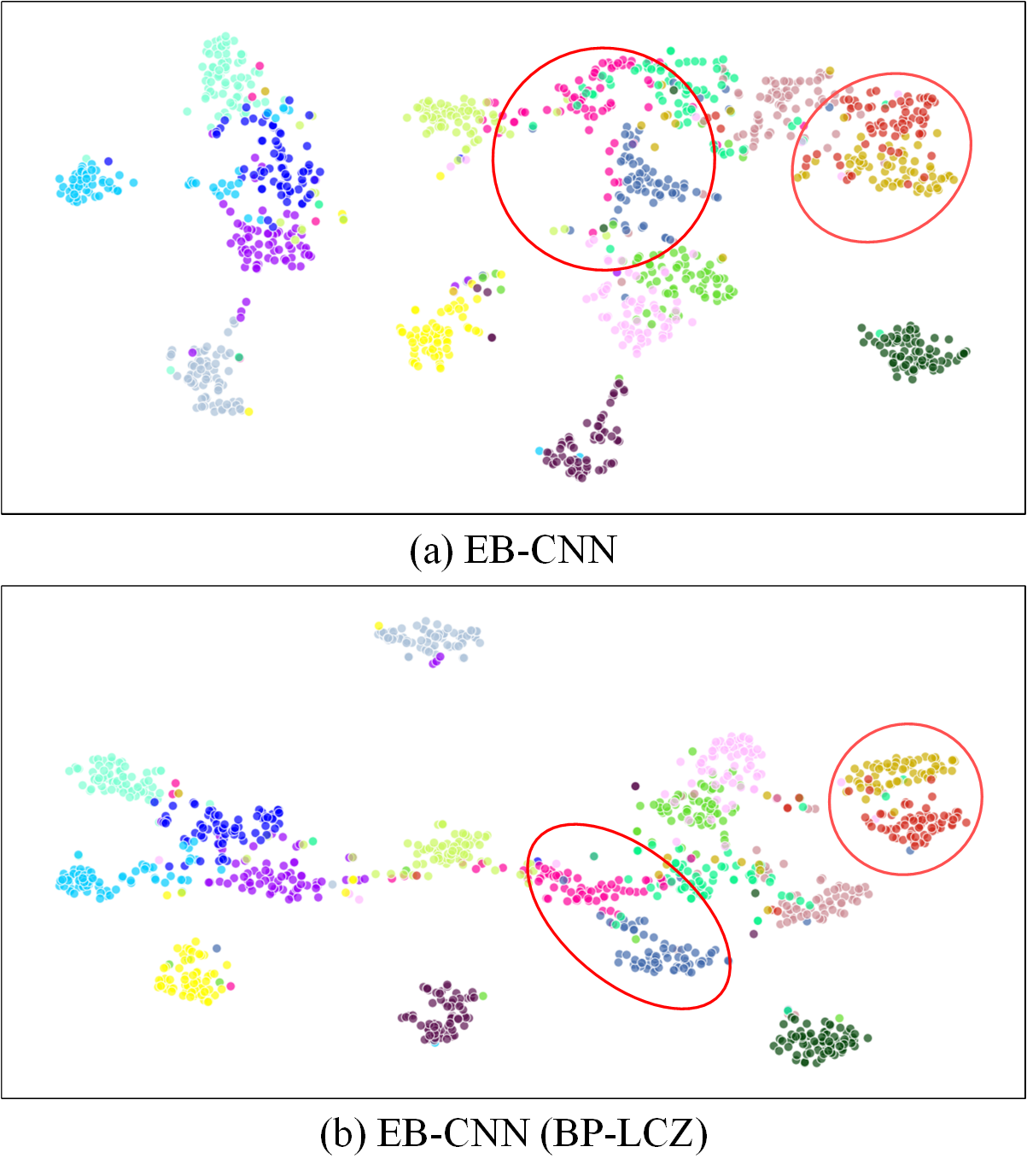}
    \caption{The t-SNE visualization of EB-CNN and EB-CNN (BP-LCZ).}
    \label{fig:tsne}
    \vspace{-6mm}
\end{figure}

As depicted in Fig. \ref{fig:confusion}, built types (e.g., compact middle-rise) and land cover types (e.g., dense tree) exhibit more misclassification within their respective types. This observation indicates that the distinction between built-type and land cover-type is more pronounced than the distinction within each type of categories. 
In addition, we randomly selected 70 samples per class for t-SNE visualization, where Fig. \ref{fig:tsne} (a) refers to EB-CNN and Fig. \ref{fig:tsne} (b) represents EB-CNN (BP-LCZ). Our strategy can better distinguish the categories in the red circles and the same category is more concentrated, indicating the effectiveness of our method.

\begin{table}[!htbp]
  \centering
  \caption{Results of ablation study.}
  \vspace{+1mm}
  \begin{tabular}{cccc}
    \toprule
    BGP & MSM & OA (\%) & Kappa (\%)\\
    \midrule
    \ding{55} & \ding{55} & 79.29          & 77.39 \\
    \ding{51} & \ding{55} & 86.15          & 84.81 \\
    \ding{51} & \ding{51} & \textbf{86.69} & \textbf{85.40}\\
    \bottomrule
  \end{tabular}
  \label{tab:oa}
  \vspace{-2mm}
\end{table}
\subsubsection{Ablation Study}
\label{sssec:ablation}
To validate the effectiveness of each component of our method, we conducted ablation experiments using EB-CNN as the baseline. As presented in Table \ref{tab:oa}, the introduction of the BGP strategy led to improvements of 6.86\% in OA and 7.42\% in the Kappa coefficient compared to the baseline, indicating that band prompting effectively exploits band-specific and semantic information. Furthermore, the introduction of the MSM-based training strategy contributed an additional increase of 0.54\% in OA and 0.59\% in the Kappa coefficient, demonstrating that the multivariate supervised matrix effectively mitigates the impact of negative samples, thereby reducing their potential to mislead the model.

\section{Conclusion}
\label{sec:conclusion}
In this paper, we presented a novel band prompting based SAR and multi-spectral data fusion framework, namely BP-LCZ, for LCZ classification. In which, a BGP strategy was introduced to facilitate SAR and multi-spectral fusion by extracting semantic information from different bands more efficiently through textual prompts and images alignment. In addition, a MSM strategy was proposed to alleviate the positive and negative sample confusion problem. The experimental results demonstrate the effectiveness and robustness of the proposed method. However, this work still faces the challenge of domain shift caused by geographical variations. Thus, we will try to tackle the domain shift issue to further enhance classification performance in future work.

\bibliographystyle{IEEEbib}
\bibliography{refs}

\end{document}